\documentclass[letterpaper,twocolumn,10pt]{article}
\usepackage{usenix-2020-09}
\usepackage{graphicx,booktabs,amsmath,amssymb,xcolor,array}
\pdftrailerid{}

\graphicspath{{./}{figs/}}
\newcommand{\AlphaHi}{1.05}
\newcommand{\AlphaLo}{0.99}
\newcommand{\BatchInvRatio}{0.55}
\newcommand{\BlowFracHi}{0.8}
\newcommand{\BlowFracLo}{0.6}
\newcommand{\BlowMaxHi}{1582}
\newcommand{\BlowMaxLo}{1237}
\newcommand{\CohEightCorr}{0.0011}
\newcommand{\CohEightOutlier}{4.8}
\newcommand{\CohEightZero}{0.19}
\newcommand{\CohLastCorr}{0.0018}
\newcommand{\CohLastCorrRank}{86}
\newcommand{\CohLastOutlier}{4.7}
\newcommand{\CohLastZero}{0.18}
\newcommand{\CohN}{448}
\newcommand{\CohShrunkCorrHi}{-0.010}
\newcommand{\CohShrunkCorrLo}{-0.027}
\newcommand{\CohShrunkOwnHi}{1.32}
\newcommand{\CohShrunkOwnLo}{0.56}
\newcommand{\CohShrunkZeroHi}{65}
\newcommand{\CountGfourHi}{4}
\newcommand{\CountGfourLo}{2}

\newcommand{\CountGfoureLo}{0}
\newcommand{\CountGtwelveHi}{4}
\newcommand{\CountGtwelveLo}{3}
\newcommand{\EbotHi}{117}
\newcommand{\EbotLo}{14}
\newcommand{\EctrlHi}{0.99}
\newcommand{\EctrlLo}{0.13}
\newcommand{\EightFourKl}{29.7}
\newcommand{\ElastHi}{0.13}
\newcommand{\ElastLo}{0.02}
\newcommand{\FloorBatchEq}{99.3}
\newcommand{\FloorBatchKL}{0.000}
\newcommand{\FloorHi}{0.21}
\newcommand{\FloorLo}{0.17}
\newcommand{\FloorN}{4}
\newcommand{\FloorReHi}{0.22}
\newcommand{\FloorReLo}{0.17}
\newcommand{\FloorTfBi}{0.81}
\newcommand{\GeiBHi}{1211}
\newcommand{\GeiBLo}{7.7}
\newcommand{\GeiCHi}{2.48}
\newcommand{\GeiCLo}{1.03}
\newcommand{\GeiRC}{0.017}
\newcommand{\GeiRD}{0.96}
\newcommand{\GeiRE}{0.09}
\newcommand{\GeiRF}{0.11}
\newcommand{\GfifteenFourN}{6}
\newcommand{\GfifteenFourOk}{6}
\newcommand{\GfifteenOneN}{12}
\newcommand{\GfifteenOneOk}{10}
\newcommand{\GfifteenPairsN}{6}
\newcommand{\GfifteenPairsOk}{5}
\newcommand{\GfifteenRows}{Qwen3-4B & int8 ch. & 1.13 & 0.57 & 0.99 & 0.98 & 0.2 (249) \\
Qwen3-4B & fp8 tensor & 0.95 & 0.44 & 0.95 & 0.95 & 0.2 (271) \\
Qwen3-4B & int4 g128 & 1.04 & 0.50 & 1.10 & 1.09 & 0.4 (247) \\
OLMo-2-7B & int8 ch. & 1.02 & 1.02 & 1.03 & 1.03 & 0.0 (5) \\
OLMo-2-7B & fp8 tensor & 0.99 & 0.99 & 1.00 & 0.99 & 0.0 (6) \\
OLMo-2-7B & int4 g128 & 1.31 & 1.31 & 1.35 & 1.35 & 0.0 (5) \\}
\newcommand{\GfifteenThreeN}{4}
\newcommand{\GfifteenThreeOk}{3}
\newcommand{\GfifteenTwoN}{12}
\newcommand{\GfifteenTwoOk}{11}
\newcommand{\GroupHi}{1.01}
\newcommand{\GroupLo}{0.96}
\newcommand{\GrowCtrlHi}{1.20}
\newcommand{\GrowCtrlLo}{1.07}
\newcommand{\GrowHi}{1.23}
\newcommand{\GrowLo}{1.07}
\newcommand{\GsixteenCtrlHi}{1.013}
\newcommand{\GsixteenCtrlLo}{1.006}
\newcommand{\GsixteenFourHi}{1.35}
\newcommand{\GsixteenFourLo}{1.05}
\newcommand{\GsixteenFourMid}{1.11}

\newcommand{\GsixteenQuarticShare}{22}
\newcommand{\GsixteenSecond}{1.04}

\newcommand{\GthirtyfourAdd}{1.03}
\newcommand{\GthirtyfourColBlock}{16}
\newcommand{\GthirtyfourOnlyElem}{33}
\newcommand{\GthirtyfourOnlyMean}{1.02}
\newcommand{\GthirtyfourOnlyOne}{196}
\newcommand{\GthirtyfourOnlyTrim}{1.34}
\newcommand{\GthirtyfourPairCorr}{0.94}
\newcommand{\GthirtyfourPairShare}{91}
\newcommand{\GthirtyfourRest}{0.93}
\newcommand{\GthirtyfourRestOne}{27}
\newcommand{\GthirtyfourRowBlock}{1.00}
\newcommand{\GthirtyfourSrMeanHi}{1.00}
\newcommand{\GthirtyfourSrMeanLo}{0.93}
\newcommand{\GthirtyfourSrOwnHi}{0.91}
\newcommand{\GthirtyfourSrOwnLo}{0.87}
\newcommand{\GthirtyfourSrTrimHi}{1.23}
\newcommand{\GthirtyfourSrTrimLo}{1.18}
\newcommand{\GthirtyfourTailK}{8}
\newcommand{\GthirtyfourTailQ}{29}
\newcommand{\GthirtyfourTwoCols}{96}
\newcommand{\GthirtyfourVerdict}{mixed}
\newcommand{\GthirtytwoBlockElem}{3.63}
\newcommand{\GthirtytwoBlockElemFp}{1.00}
\newcommand{\GthirtytwoBlockElemInt}{1.08}
\newcommand{\GthirtytwoBlockElemSmallHi}{1.05}
\newcommand{\GthirtytwoBlockElemSmallLo}{1.00}
\newcommand{\GthirtytwoBlockOne}{1.20}
\newcommand{\GthirtytwoBlockOneTrim}{1.46}
\newcommand{\GthirtytwoBlockQ}{0.93}
\newcommand{\GthirtytwoBlockQTrim}{1.13}
\newcommand{\GthirtytwoCK}{-0.18}
\newcommand{\GthirtytwoDiagQ}{3.4}
\newcommand{\GthirtytwoDiagQTrim}{4.0}
\newcommand{\GthirtytwoFourKl}{193}
\newcommand{\GthirtytwoFourN}{3}
\newcommand{\GthirtytwoFourOk}{3}
\newcommand{\GthirtytwoFourTrim}{5.18}
\newcommand{\GthirtytwoFourTrimBf}{5.17}
\newcommand{\GthirtytwoFpTrim}{0.86}
\newcommand{\GthirtytwoFullBlock}{0.93}
\newcommand{\GthirtytwoFullBlockHi}{1.11}
\newcommand{\GthirtytwoFullBlockLo}{0.89}
\newcommand{\GthirtytwoHybQ}{0.94}
\newcommand{\GthirtytwoHybQTrim}{1.15}
\newcommand{\GthirtytwoIntTrim}{0.99}
\newcommand{\GthirtytwoN}{256}
\newcommand{\GthirtytwoOneN}{6}
\newcommand{\GthirtytwoOneOk}{4}
\newcommand{\GthirtytwoRestBlockElem}{1.02}
\newcommand{\GthirtytwoSlope}{2.18}
\newcommand{\GthirtytwoSpBlockFour}{0.25}
\newcommand{\GthirtytwoSpFour}{0.09}

\newcommand{\GthirtytwoThreeN}{2}
\newcommand{\GthirtytwoThreeOk}{1}
\newcommand{\GthirtytwoTopBlockShare}{74}
\newcommand{\GthirtytwoTopElemShare}{8}
\newcommand{\GthirtytwoTopExcess}{99}
\newcommand{\GthirtytwoTopOwn}{32}
\newcommand{\GthirtytwoTopOwnInt}{0.91}
\newcommand{\GthirtytwoTopSeqShare}{29}
\newcommand{\GthirtytwoTopSign}{0.07}
\newcommand{\GthirtytwoTwoN}{6}
\newcommand{\GthirtytwoTwoOk}{6}
\newcommand{\HillKHi}{12}
\newcommand{\HillKLo}{2.5}
\newcommand{\HillQHi}{2.4}
\newcommand{\HillQLo}{1.7}
\newcommand{\HpropN}{2}
\newcommand{\HwgradN}{0}
\newcommand{\JumpCtrlHi}{1.13}
\newcommand{\JumpHi}{11.8}
\newcommand{\JumpLo}{4.4}
\newcommand{\LastDownOwnSmallHi}{1.19}
\newcommand{\LastDownOwnSmallLo}{0.86}
\newcommand{\LocalCtrlHi}{1.04}
\newcommand{\LocalCtrlLo}{0.93}
\newcommand{\LocalLastHi}{1.01}
\newcommand{\LocalLastLo}{0.94}
\newcommand{\LocalShrunkRow}{11}
\newcommand{\MassiveTokOk}{0}
\newcommand{\MaxGfour}{27.2}
\newcommand{\MaxGfoure}{4.5}
\newcommand{\MaxGtwelve}{1581.8}

\newcommand{\MeasOverGfoureFp}{1.00}
\newcommand{\MeasOverGfoureInt}{1.02}
\newcommand{\MedianRatio}{1.00}
\newcommand{\MoeBlockElemHi}{1.03}
\newcommand{\MoeBlockElemLo}{1.01}
\newcommand{\MoeFlipHi}{11.7}
\newcommand{\MoeFlipLo}{6.6}
\newcommand{\MoeFourSlope}{1.65}
\newcommand{\MoeFourTrim}{1.09}
\newcommand{\MoeFpTrim}{2.1}
\newcommand{\MoeIntFlipSlope}{0.41}
\newcommand{\MoeIntQuarter}{27}
\newcommand{\MoeIntSlope}{0.63}
\newcommand{\MoeIntTrim}{4.1}
\newcommand{\MoeN}{256}
\newcommand{\MoeRouteDisagree}{12}
\newcommand{\Molmo}{171}
\newcommand{\Mqwen}{34{,}656}
\newcommand{\Mratio}{203}
\newcommand{\NAllSeq}{1{,}280}
\newcommand{\NGfour}{256}
\newcommand{\NGfoure}{512}
\newcommand{\NGtwelve}{512}
\newcommand{\NOverThousand}{1}
\newcommand{\NTrimEight}{18}
\newcommand{\OlmoBlowMax}{6}
\newcommand{\OlmoFourKl}{20.6}
\newcommand{\OlmoFourMeanHi}{1.31}

\newcommand{\OlmoFourTrim}{1.35}
\newcommand{\OlmoMeanTrimGap}{0.04}
\newcommand{\OlmoRelFour}{1.28}

\newcommand{\PredOverGfoureFour}{4.52}

\newcommand{\PredOverGfoureInt}{4.04}
\newcommand{\QfourBlowHi}{0.4}
\newcommand{\QfourBlowLo}{0.2}
\newcommand{\QfourBlowMax}{271}
\newcommand{\QfourMeanBfHi}{0.57}
\newcommand{\QfourMeanBfLo}{0.44}
\newcommand{\QfourMeanFpHi}{1.13}
\newcommand{\QfourMeanFpLo}{0.95}
\newcommand{\QfourMeanOverTrim}{1.13}
\newcommand{\QfourRelHi}{0.92}
\newcommand{\QfourRelLo}{0.84}
\newcommand{\QfourTrimBfHi}{1.09}
\newcommand{\QfourTrimBfLo}{0.95}

\newcommand{\QfourTrimIntFp}{0.99}

\newcommand{\RbfEFour}{0.97}
\newcommand{\RbfEFp}{1.01}
\newcommand{\RbfEInt}{1.15}

\newcommand{\RexpThreeFour}{0.22}
\newcommand{\RexpThreeFp}{0.23}
\newcommand{\RexpThreeInt}{0.29}
\newcommand{\RfpEFour}{0.70}

\newcommand{\RfpEInt}{0.66}
\newcommand{\RfpTFour}{0.58}

\newcommand{\RfpTInt}{0.56}
\newcommand{\RgFourHi}{0.88}
\newcommand{\RgFourLo}{0.85}
\newcommand{\RpreFour}{0.98}
\newcommand{\RpreFp}{0.98}
\newcommand{\RpreInt}{1.16}
\newcommand{\RruleFour}{1.03}
\newcommand{\RruleFp}{1.02}
\newcommand{\RruleInt}{0.86}
\newcommand{\SignNullP}{1.1\times10^{-6}}
\newcommand{\SrKl}{2.25}
\newcommand{\SrPred}{2.28}
\newcommand{\TailSignAgree}{30}
\newcommand{\TailSignN}{32}
\newcommand{\TokShareHi}{58}
\newcommand{\TokShareLo}{28}
\newcommand{\TopFiveShare}{24}
\newcommand{\TrimEightHi}{1.08}
\newcommand{\TrimEightLo}{0.95}
\newcommand{\TrimOverMeanHi}{1.22}
\newcommand{\TrimOverMeanLo}{1.09}

\newcommand{\ScrAcMisAgreeHi}{94}
\newcommand{\ScrAcMisAgreeLo}{81}
\newcommand{\ScrAcQAgreeHi}{94}
\newcommand{\ScrAcQAgreeLo}{88}
\newcommand{\ScrAcQBitsBlk}{4.14}
\newcommand{\ScrAcQBitsDelta}{0.02}
\newcommand{\ScrAcQBitsElm}{4.12}
\newcommand{\ScrAcQBlkPred}{33.6}

\newcommand{\ScrAcQCross}{55}
\newcommand{\ScrAcQElemUFour}{36.4}
\newcommand{\ScrAcQOver}{3.5}
\newcommand{\ScrBandHi}{1.50}
\newcommand{\ScrBandLo}{0.75}
\newcommand{\ScrBitsMOne}{6.55}
\newcommand{\ScrBitsMThree}{7.64}
\newcommand{\ScrBitsMTwo}{7.28}
\newcommand{\ScrBitsUEight}{8.00}
\newcommand{\ScrBitsUFour}{4.12}
\newcommand{\ScrBitsUFp}{8.00}
\newcommand{\ScrBudgetMOne}{5.41}
\newcommand{\ScrBudgetMThree}{1.35}
\newcommand{\ScrBudgetMTwo}{2.71}
\newcommand{\ScrCalibN}{17}
\newcommand{\ScrCalibUniHi}{1.46}
\newcommand{\ScrCalibUniLo}{0.87}
\newcommand{\ScrCurseHi}{1.022}
\newcommand{\ScrCurseLo}{1.015}
\newcommand{\ScrCurseSmokeHi}{1.44}
\newcommand{\ScrCurseSmokeLo}{1.16}
\newcommand{\ScrFisherMin}{8.0}
\newcommand{\ScrFisherSecPerSeq}{1.88}

\newcommand{\ScrGFullKlAlphaOne}{193.2}

\newcommand{\ScrGOneAlphaQuarter}{1.34}
\newcommand{\ScrGOneKlAlphaOne}{195.8}
\newcommand{\ScrGOneMeanAlphaOne}{1.64}
\newcommand{\ScrGOneMeanAlphaQuarter}{1.02}
\newcommand{\ScrGRestAlphaOne}{0.77}
\newcommand{\ScrGRestKlAlphaOne}{27.4}
\newcommand{\ScrGSrDrop}{15}
\newcommand{\ScrGSrMeanHi}{1.00}
\newcommand{\ScrGSrMeanLo}{0.93}
\newcommand{\ScrGSrRatioHi}{1.23}
\newcommand{\ScrGSrRatioLo}{1.18}
\newcommand{\ScrGTwoColShare}{96}
\newcommand{\ScrGVerdict}{MIXED}
\newcommand{\ScrGainMOne}{8}
\newcommand{\ScrGainMThree}{4}
\newcommand{\ScrGainMTwo}{5}
\newcommand{\ScrGainUEight}{3}
\newcommand{\ScrGainUFour}{17}
\newcommand{\ScrGpuHAll}{1.06}
\newcommand{\ScrGpuHMain}{0.77}
\newcommand{\ScrHits}{6}
\newcommand{\ScrHitsDraft}{6}
\newcommand{\ScrHitsElem}{6}
\newcommand{\ScrHitsElemGfifteen}{5}
\newcommand{\ScrHitsGfifteen}{6}
\newcommand{\ScrKlMOne}{3.58}
\newcommand{\ScrKlMThree}{0.891}
\newcommand{\ScrKlMTwo}{1.74}
\newcommand{\ScrKlMin}{8.7}
\newcommand{\ScrKlUEight}{0.127}
\newcommand{\ScrKlUFour}{12.1}
\newcommand{\ScrKlUFp}{0.473}
\newcommand{\ScrLamBf}{74.8}

\newcommand{\ScrLamMOne}{74.6}
\newcommand{\ScrLamMThree}{74.7}
\newcommand{\ScrLamMTwo}{74.5}
\newcommand{\ScrLamN}{5153}
\newcommand{\ScrLamUEight}{75.0}
\newcommand{\ScrLamUFour}{73.2}
\newcommand{\ScrLayers}{32}
\newcommand{\ScrMOneIntFourLayers}{12}

\newcommand{\ScrMaxOwnFour}{2.6}
\newcommand{\ScrMixRatioHi}{1.08}
\newcommand{\ScrMixRatioLo}{1.06}
\newcommand{\ScrModel}{Mistral-7B-v0.3}
\newcommand{\ScrN}{256}
\newcommand{\ScrParamsB}{7.0}
\newcommand{\ScrPredMOne}{3.31}
\newcommand{\ScrPredMThree}{0.836}
\newcommand{\ScrPredMTwo}{1.64}
\newcommand{\ScrPredUEight}{0.115}
\newcommand{\ScrPredUFour}{10.8}
\newcommand{\ScrPredUFp}{0.459}

\newcommand{\ScrRatioElemUFourX}{1.155}

\newcommand{\ScrRatioHi}{1.12}
\newcommand{\ScrRatioLo}{1.03}
\newcommand{\ScrRatioMOne}{1.08}
\newcommand{\ScrRatioMThree}{1.07}
\newcommand{\ScrRatioMTwo}{1.06}
\newcommand{\ScrRatioUEight}{1.11}
\newcommand{\ScrRatioUFour}{1.12}
\newcommand{\ScrRatioUFp}{1.03}
\newcommand{\ScrRetroCells}{19}
\newcommand{\ScrRetroMaxOwnOther}{6.4}
\newcommand{\ScrSmokeRatioHi}{1.41}
\newcommand{\ScrSmokeRatioLo}{1.20}
\newcommand{\ScrTauExcPct}{5}
\newcommand{\ScrTauOwn}{10}
\newcommand{\ScrTensors}{224}
\newcommand{\ScrTpsBaseBfHi}{7.3}
\newcommand{\ScrTpsBaseBfLo}{3.7}
\newcommand{\ScrTpsBaseEightHi}{4.5}
\newcommand{\ScrTpsBaseEightLo}{0.7}
\newcommand{\ScrTpsBf}{4025}

\newcommand{\ScrTpsErrMax}{0.4}
\newcommand{\ScrTpsMOne}{4343}
\newcommand{\ScrTpsMThree}{4180}
\newcommand{\ScrTpsMTwo}{4226}
\newcommand{\ScrTpsPredMOne}{4343}
\newcommand{\ScrTpsPredMThree}{4196}
\newcommand{\ScrTpsPredMTwo}{4244}
\newcommand{\ScrTpsRepSpread}{0.6}
\newcommand{\ScrTpsUEight}{4150}
\newcommand{\ScrTpsUFour}{4709}
\newcommand{\ScrVerdict}{SCREENER\_PASS}

\newcommand{\ScrVklRatioMTwo}{0.87}

\newcommand{\ScrVklRatioUFour}{0.99}

\begin{document}
\date{}
\title{\Large \bf How Far Does Second-Order Linear Response Predict Quantization Mismatch?\\ Pre-Registered Measurements on Five Language Models}
\author{{\rm Dezhen Lu}\\ University of T\"ubingen \and {\rm Zixiang Ni}\\ Xi'an Jiaotong University}
\maketitle

\begin{abstract}
	Serving systems choose a weight format for each model and deployment.
	A low-cost screen predicts the resulting KL divergence from the unquantized
	model's Fisher information using $\tfrac{1}{2}\Delta^\top F\Delta$, the
	expansion underlying Hessian-aware and second-order post-training quantizers.
	We assess the reliability of this screen in GPU batches with channel-wise
	int8, per-tensor fp8 and group-wise int4 on Qwen3-8B, Qwen3-4B, OLMo-2-7B,
	Qwen3-32B and Qwen3-30B-A3B.
	Relative predictions hold across a 16-fold range of perturbation scales,
	int4 group sizes and rounding modes, while absolute predictions are heavy-tailed.
	A trimmed average brings 10 of 12 cells on Qwen3-4B and OLMo-2 within
	15\% of measured KL.
	The MoE model exhibits nonquadratic response at small perturbations,
	with int8 KL scaling as $\alpha^{0.63}$.
	Computing per-sequence gradients in bf16 can cause fourfold prediction errors.
	We trace these errors to arithmetic in the lower half of the backward pass
	and eliminate them by using fp32 there.
	Batch-invariant kernels run at 0.55 times the throughput, while changes
	in reduction order introduce a floor of 0.17--0.22 nats per sequence.
	We build a screener with one streamed fp32 Fisher pass, trimmed tensor-block
	predictions with a band, a flag for aligned tensors and per-layer format allocation.
	On the unseen Mistral-7B-v0.3 family, all six predictions fall within the band,
	all three KL budgets are met, and vLLM throughput is predicted within 0.4\%.
\end{abstract}

\section{Introduction}

An LLM serving stack now picks a weight format for every model and deployment: int8 or fp8 where quality matters most, int4 with a
group size where memory binds, block-scaled 4-bit formats~\cite{rouhani2023mx} and, on new GPUs, stochastic instead of nearest
rounding~\cite{gupta2015limited}. Measuring task quality for every combination is slow, so a screen that predicts the damage from the
unquantized model is attractive. The standard one is second order: the KL divergence between a model and its quantized copy is
approximately $\tfrac12\Delta^\top F\Delta$, with $F$ the Fisher information and $\Delta$ the weight perturbation. It costs one
forward and backward pass per calibration sequence, after which every candidate format is a dot product per tensor; Hessian-aware bit
allocation~\cite{dong2019hawq} and second-order post-training quantizers~\cite{frantar2022obc,frantar2023gptq,nagel2020adaround}
build on the same expansion.

A screen is used in two ways. It \emph{ranks} candidates on a fixed calibration set, and it gives an \emph{absolute} number that is
compared with a quality budget. We test both, together with the implementation choices that decide whether the number can be trusted:
the precision of the per-sequence gradients, the estimator over sequences, and the noise floor of the engine that measures the KL. The
question is one of linear response: the quantized model is a perturbed copy, and the KL is its quadratic response in the Fisher metric
(\S\ref{sec:theory}). Every decision rule was frozen---predictions, scorers and file checksums---before its GPU run, and the models
added after Qwen3-8B played no part in developing the analysis. The ledger in the supplementary material lists every batch with its frozen prediction
and outcome.

\paragraph{Findings.}
(1) Relative predictions hold: across perturbation scale, int4 group size and rounding mode (\S\ref{sec:rel}).
(2) The absolute scale is heavy-tailed: the plain average over sequences is set by a few of them (\S\ref{sec:heavy}); a trimmed
average fixes most cells on new models (\S\ref{sec:trim}) but not int4 at 32B, where the diagonal estimator misses rounding error that is coherent along the gradient in one tensor (\S\ref{sec:big}).
(3) bf16 per-sequence gradients can be off by $4\times$ through a handful of sequences whose backward pass is numerically unstable in
the lower layers; we locate where the error is made and give a recipe (\S\ref{sec:bf}).
(4) The engine's run-to-run floor, and what removing it costs in throughput (\S\ref{sec:floor}).
(5) \S\ref{sec:guide} turns these into a checklist for a quantization screen, and \S\ref{sec:screener} builds the screen: one
streamed Fisher pass, a band, a flag for aligned tensors and a per-layer allocator that writes vLLM checkpoints. On an unseen
model family it met its pre-registered criteria for KL, budgets and throughput.

\section{Setup}

\paragraph{Model and schemes.} Qwen3-8B; weight quantization int8 per channel, fp8 per tensor, int4 with group size 128 (and 32, 64,
256; nearest and stochastic rounding). $\Delta=\alpha(\hat W-W)$ with $\alpha\in\{1/8,1/4,1/2,1\}$ to test the expansion.
Two further models, chosen before any run on them: Qwen3-4B (same family and tokenizer, tied input and output embeddings) and
OLMo-2-1124-7B (a different family), each with the three main schemes at $\alpha=1$; and Qwen3-32B (four times larger), with an
$\alpha$ scan of int4 in $\{1/4,1/2,1\}$.
\paragraph{Measured mismatch.} Token-level KL between the base and the perturbed model on sampled continuations of 256 tokens,
averaged per sequence, then over sequences.
\paragraph{Prediction.} Per sequence, $q=\tfrac12\sum_i g_i^2\Delta_i^2$ over all weights $i$, with $g$ the gradient of the sequence
log-likelihood: a one-sample estimate of the diagonal part of $\tfrac12\Delta^\top F\Delta$. The continuations are sampled from the
model, so $F$ is the Fisher, not the empirical Fisher~\cite{kunstner2019limitations}. The same pass also records each tensor's
projection $d_\ell=g_\ell\cdot\Delta_\ell$, from which the tensor-block form $\tfrac12\sum_\ell d_\ell^2$ and the full one-sample form
$\tfrac12(\sum_\ell d_\ell)^2$ follow (\S\ref{sec:big}). The prediction is the mean of $q$, computed
with fp32 and with bf16 backward passes. $R$ = measured KL / prediction.
\paragraph{Sample sets.} A: 256 prompts, samples from an HF fp32 sampler. B: 512 prompts (the original pipeline's prompt rule), vLLM
samples. C: the same 512 prompts, samples produced by the original pipeline's sampler. The two further models use set B's rules (for
OLMo-2 the prompts are cut by the same rule with its own tokenizer). All runs on GH200; versions pinned.

\section{Theory: linear response in the Fisher metric}\label{sec:theory}

The expansion $\mathrm{KL}\approx\tfrac12\Delta^\top F\Delta$ is the quadratic response of the model's output distribution to a
perturbation of its parameters, with the Fisher information as the metric. Four things can break it, and the paper tests each. First,
the response is a sum over sequences of a quadratic form, and that sum can be dominated by rare configurations: a heavy tail makes the
mean not self-averaging, while ratios on a fixed sample set remain accurate (\S\ref{sec:rel}, \S\ref{sec:heavy}). Second, the adjoint
dynamics that produce the per-sequence gradients can be unstable in reduced precision and amplify a rounding error on its way back through the layers; the blow-up is then a
property of the backward pass, not of the model, and in our data a single layer supplies the step that separates blow-ups from controls (\S\ref{sec:bf}). Third, the perturbation
can leave the quadratic regime, which an $\alpha$ scan detects as a higher-order term (\S\ref{sec:trim}). Fourth, the diagonal
estimator treats the rounding errors of a tensor as sign-incoherent along the gradient; when they are coherent it under-predicts by a
factor that does not close as $\alpha\to0$ (\S\ref{sec:big}). Rounding also supplies a
quenched/annealed pair: round-to-nearest is a fixed (quenched) error for a given input and stochastic rounding an annealed one, and the
variance term of the expansion predicts their ratio. The run-to-run floor of a bf16 engine is this layer's thermal noise: any change of
reduction order realises a new rounding configuration of the same size (\S\ref{sec:floor}).

\section{Relative predictions hold}\label{sec:rel}

$R$ changed by a factor \AlphaLo--\AlphaHi{} between $\alpha=1/8$ and $\alpha=1$, so the quadratic term dominates. For int4 group
sizes 32, 64 and 256, calibrating on group size 128 predicted the measured KL within \GroupLo--\GroupHi, and KL was monotone in group
size. Stochastic rounding raised the KL by \SrKl$\times$ against a predicted \SrPred$\times$. On the two new models, three of four
pre-registered scheme ratios were within $\pm25\%$ (Section~\ref{sec:trim}); the miss is int4 against int8 on OLMo-2 (\OlmoRelFour).

\section{The absolute scale is heavy-tailed}\label{sec:heavy}

With fp32 gradients, $R$ ranged from \RgFourLo--\RgFourHi{} on set A to \RfpEInt--\RfpEFour{} on set B and \RfpTInt--\RfpTFour{} on set
C. The per-sequence quadratic form is heavy-tailed: the top 5\% of sequences carry up to \TopFiveShare\% of the prediction. Post hoc, a Hill
estimate on the top 10\% gives the per-sequence $q$ a tail index of \HillQLo--\HillQHi{} on Qwen3-8B and Qwen3-32B, at the edge of
infinite variance, and the per-sequence KL \HillKLo--\HillKHi. Because $R_\text{trim}/R_\text{mean}=(1-s_K)/(1-s_q)$ exactly, with $s$ the
share of a quantity carried by its top 5\%, the heavier tail of $q$ puts the trimmed ratio \TrimOverMeanLo--\TrimOverMeanHi$\times$ above the
ratio of means on those two models, and the sign of the difference follows the two tail indices in \TailSignAgree{} of \TailSignN{} cells
(four models, three schemes, both forms of $q$). An absolute
prediction therefore carries a sample-set error of tens of percent even when relative comparisons on the same samples are accurate.
The size of this bias depends on the model: on Qwen3-4B the fp32 ratio was \QfourMeanFpLo--\QfourMeanFpHi, and on OLMo-2 the plain and
trimmed averages nearly coincide (Table~\ref{tab:g15}).

\section{bf16 backward passes can be off by $4\times$}\label{sec:bf}

\paragraph{Reproduction.} An earlier experiment reported that the prediction over-estimated KL by about $4.2\times$ uniformly across
schemes. Running that experiment's own code unchanged on GH200 reproduced it ($R=\RexpThreeInt, \RexpThreeFp, \RexpThreeFour$ for int8,
fp8, int4), with the measured KL matching our independent pipeline (\MeasOverGfoureInt--\MeasOverGfoureFp$\times$) and the prediction
\PredOverGfoureInt--\PredOverGfoureFour$\times$ higher. A different transformers version gave the same result. Our own code on the
\emph{same samples} (set C) gave the same numbers to five or six significant figures; on set B it gave $R=\RbfEInt, \RbfEFp, \RbfEFour$.
The difference is in the samples.

\begin{figure}[t]\centering
\includegraphics[width=\columnwidth]{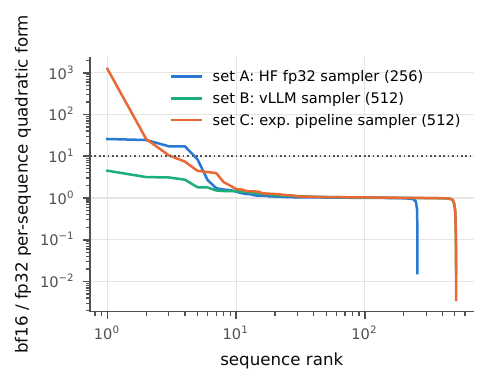}
\caption{Per-sequence ratio of the bf16 to the fp32 quadratic form (int8), sorted. Dotted: the post-hoc threshold 10.}\label{fig:blow}
\end{figure}

\begin{figure}[t]\centering
\includegraphics[width=\columnwidth]{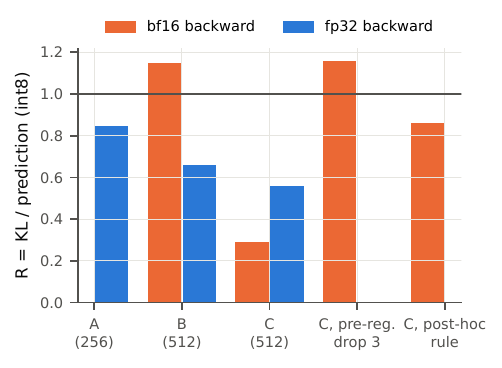}
\caption{$R$ by sample set and gradient precision (int8). On set C the bf16 estimate is off by $3$--$4\times$ until the blown-up
sequences are removed.}\label{fig:R}
\end{figure}

\paragraph{Mechanism.} For most sequences the bf16 and fp32 quadratic forms agree (median ratio \MedianRatio). On set C, a few
sequences have early-layer gradients that blow up element-wise in bf16---one reaches \BlowMaxLo--\BlowMaxHi$\times$ its fp32 value---and
the squares carry them into the mean, while the linear term $\sum g\cdot\Delta$ is barely affected (Fig.~\ref{fig:blow}).
\begin{itemize}
\item Pre-registered: dropping the three sequences flagged before the run gives $R=\RpreInt, \RpreFp, \RpreFour$.
\item Post hoc: dropping every sequence with bf16/fp32 $>10$ (\BlowFracLo--\BlowFracHi\% of the set; threshold chosen after seeing the
  data) gives $R=\RruleInt, \RruleFp, \RruleFour$ (Fig.~\ref{fig:R}).
\end{itemize}
\paragraph{How often.} Counting sequences with bf16/fp32 $>10$ per scheme: set A \CountGfourLo--\CountGfourHi{} of \NGfour{}
(max \MaxGfour$\times$), set B \CountGfoureLo{} of \NGfoure{} (max \MaxGfoure$\times$), set C \CountGtwelveLo--\CountGtwelveHi{} of
\NGtwelve{} (max \MaxGtwelve$\times$). Only \NOverThousand{} of the \NAllSeq{} sequences exceeded $1000\times$. The original run's own per-sequence
data were not inspected; we attribute it by reproduction.

\paragraph{Where the error enters.} We back-propagated the four blown-up sequences of set C and two controls once in fp32 and once in
bf16 and compared the gradients with respect to every hidden state (pre-registered). The bf16 gradients reproduced the blow-ups bit for
bit. The relative error of the hidden-state gradient is \ElastLo--\ElastHi{} at the last layer and \EbotLo--\EbotHi{} at the output
of layer 1 (layers counted from 0), so the error lives in the residual-stream gradient and grows toward the input; it is not confined to the weight-gradient products
(that hypothesis held for \HwgradN{} of 4; the stricter form requiring $\le0.1$ at the last layer held for \HpropN{} of 4). Post hoc, the
error jumps by \JumpLo--\JumpHi$\times$ in the backward step through layer 15 and grows by \GrowLo--\GrowHi$\times$ per layer below it.
It is not concentrated on one token (the largest token carries \TokShareLo--\TokShareHi\% of the squared error) and not on the token with
the largest forward activation (\MassiveTokOk{} of 4), which refutes our guess that massive activations inject it. The two controls show
the same growth below layer 15 (\GrowCtrlLo--\GrowCtrlHi$\times$ per layer) on a smaller scale (relative error \EctrlLo{} and \EctrlHi{} at the
output of layer 1) but no jump there (at most \JumpCtrlHi$\times$): what separates the blow-ups is one layer, not a faster per-layer growth. The largest activation is about \Mqwen{} times the median in Qwen3-8B and \Molmo{} times in OLMo-2 (ratio \Mratio),
consistent with---but not shown to cause---the absence of blow-ups in OLMo-2.

A frozen follow-up separated where the error is made from where it grows. With the backward of layers 0--15 in fp32 and the rest in
bf16, the error at the bottom fell to \GeiRC{} of the all-bf16 value (median over the four sequences); with the upper layers in fp32 it
stayed at \GeiRD. The quadratic form, \GeiBLo--\GeiBHi$\times$ its fp32 value in all-bf16, came to \GeiCLo--\GeiCHi$\times$ with the lower
layers in fp32. The blow-ups are therefore made by bf16 arithmetic in the lower half; the backward Jacobian also amplifies the small
error that arrives from above, but two orders of magnitude less. Which block is responsible depends on the sequence (attention or MLP
in fp32 alone leave \GeiRE{} and \GeiRF{} of the error at the median, but each fails on a different sequence), so a practical recipe keeps
the whole lower half of the backward in fp32.

\section{A trimmed average holds on two new models, except int4 on OLMo-2}\label{sec:trim}

The heavy tail suggests a robust average. Let $R_{\text{trim}}=\operatorname{trim}(\mathrm{KL})/\operatorname{trim}(q)$, where trim
drops the largest 5\% of a series (KL and $q$ are trimmed separately); it estimates the KL of a typical sequence rather than a mean set by
a few. On Qwen3-8B sets A--C it gave \TrimEightLo--\TrimEightHi{} in all \NTrimEight{} cells (three sets, three schemes, two gradient precisions).
Because we chose it after seeing those data, we pre-registered four criteria for the two new models before running them (Table~\ref{tab:g15}):
\begin{itemize}
\item P1, $R_{\text{trim}}\in[0.85,1.15]$: \GfifteenOneOk/\GfifteenOneN. Both misses are OLMo-2 int4 (\OlmoFourTrim). The fp32 and bf16
  cells of a model and scheme share the measured KL, so by model and scheme this is \GfifteenPairsOk/\GfifteenPairsN.
\item P2, $R_{\text{mean}}<R_{\text{trim}}$: \GfifteenTwoOk/\GfifteenTwoN. The exception is Qwen3-4B int8 with fp32 gradients
  (\QfourMeanOverTrim{} against \QfourTrimIntFp). On OLMo-2 the margins are at most \OlmoMeanTrimGap: without blow-ups the tail of $q$ is barely
  heavier than that of the KL.
\item P3, scheme ratios within $\pm25\%$: \GfifteenThreeOk/\GfifteenThreeN{} (Qwen3-4B \QfourRelLo--\QfourRelHi; OLMo-2 int4/int8
  \OlmoRelFour).
\item P4, at most 1\% of sequences with bf16/fp32 $>10$: \GfifteenFourOk/\GfifteenFourN. On Qwen3-4B \QfourBlowLo--\QfourBlowHi\% of
  sequences blow up (up to \QfourBlowMax$\times$), enough to pull the bf16 plain average down to \QfourMeanBfLo--\QfourMeanBfHi;
  trimming restores \QfourTrimBfLo--\QfourTrimBfHi. OLMo-2 has none (largest ratio \OlmoBlowMax).
\end{itemize}
The OLMo-2 int4 miss appears under both estimators and both gradient precisions ($R_{\text{mean}}=\OlmoFourMeanHi$), so it is neither a
tail nor a bf16 effect. It could be a higher-order term, which would vanish as $\alpha\to0$, or a bias of the Fisher estimate along
the int4 direction, which would not. A pre-registered $\alpha$ scan on OLMo-2 (same samples and Fisher arrays; int8 as control) decided
for the first: $R_{\text{trim}}$ for int4 rose monotonically from \GsixteenFourLo{} at $\alpha=1/8$ through \GsixteenFourMid{} at
$1/2$ to \GsixteenFourHi{} at 1, while int8 stayed flat (\GsixteenCtrlLo--\GsixteenCtrlHi). Fitting
$\mathrm{KL}(\alpha)=a\alpha^2+b\alpha^3+c\alpha^4$ gives a second-order coefficient \GsixteenSecond{} times the Fisher prediction
and a quartic term that supplies \GsixteenQuarticShare\% of the KL at $\alpha=1$. Size alone would not have told: OLMo-2's int4 KL
(\OlmoFourKl{} per sequence) is smaller than Qwen3-8B's (\EightFourKl), where the ratio did not depend on $\alpha$.

\begin{table}[t]\centering\footnotesize\setlength{\tabcolsep}{3.5pt}
\caption{Measured KL over prediction on Qwen3-4B and OLMo-2, by estimator and gradient precision; last column: share of sequences with
bf16/fp32 $>10$ (largest ratio). Pre-registered range for $R_{\text{trim}}$: $[0.85, 1.15]$.}\label{tab:g15}
\begin{tabular}{llrrrrr}
\toprule
Model & Scheme & \multicolumn{2}{c}{$R_{\text{mean}}$} & \multicolumn{2}{c}{$R_{\text{trim}}$} & Blow-ups \\
 & & fp32 & bf16 & fp32 & bf16 & \% (max) \\
\midrule
\GfifteenRows
\bottomrule
\end{tabular}
\end{table}

\section{At 32B the diagonal form misses int4}\label{sec:big}

Our $q$ is the diagonal part of the quadratic form, so it is exact only if the rounding errors of each tensor are sign-incoherent along
the gradient. We pre-registered G15's four criteria for Qwen3-32B (\GthirtytwoN{} sequences, set B's rules, fp32 and bf16 backward;
per-tensor gradients streamed through hooks so that the fp32 model fits on three GPUs) together with an $\alpha$ scan of int4 meant as
the dense control for a planned mixture-of-experts test. int8 and fp8 held ($R_{\text{trim}}$ \GthirtytwoIntTrim{} and
\GthirtytwoFpTrim), int4 did not (\GthirtytwoFourTrim{} with fp32 and \GthirtytwoFourTrimBf{} with bf16 gradients): P1
\GthirtytwoOneOk/\GthirtytwoOneN, P2 \GthirtytwoTwoOk/\GthirtytwoTwoN, P3 \GthirtytwoThreeOk/\GthirtytwoThreeN, P4
\GthirtytwoFourOk/\GthirtytwoFourN. The control failed as well: the int4 KL grew with slope \GthirtytwoSlope{} in $\alpha$ and a two-term
fit gave $C/K_1=\GthirtytwoCK$ (bands $[1.9,2.1]$ and $|C/K_1|<0.05$), so by the frozen rule the mixture-of-experts test that depends on
this control is inconclusive whatever it measures, and the run itself stopped in its last, unrelated phase (the bf16 pipeline; the Fisher
and the fp32 KL scan were complete). We report Qwen3-30B-A3B (\MoeN{} sequences) as an observation. The pre-registered alternative,
$\mathrm{KL}=A\alpha^2+C\alpha$ with $C$ from routing flips, is classed \emph{mixed} for both schemes: for int8 the share of (layer, token)
whose expert set flips grows only as $\alpha^{\MoeIntFlipSlope}$ (\MoeFlipLo\% at $\alpha=1/4$, \MoeFlipHi\% at 1) and the KL as
$\alpha^{\MoeIntSlope}$, both sub-linear. The consequence for the expansion is the reverse of the dense miss: the smaller the perturbation, the
worse it predicts. Trimmed, int8 comes out at \MoeIntTrim$\times$ the prediction at full scale and \MoeIntQuarter$\times$ at a quarter, fp8
at \MoeFpTrim$\times$, while int4 is close (\MoeFourTrim, KL $\propto\alpha^{\MoeFourSlope}$). There is no coherence to blame
(block/diagonal \MoeBlockElemLo--\MoeBlockElemHi, full form within the random-sign range), and the fp32 and bf16 references already
disagree on \MoeRouteDisagree\% of routing decisions: on this model the quadratic form misses a non-analytic part of the response.

The miss is not OLMo-2's higher-order term: at $\alpha=1/4$ the int4 ratio is still \GthirtytwoDiagQTrim{} (\GthirtytwoDiagQ{} by the
plain mean). Post hoc, the recorded projections $d_\ell$ give the tensor-block form $\tfrac12\sum_\ell d_\ell^2$, exact within each
tensor, and the full form $\tfrac12(\sum_\ell d_\ell)^2$. For int4 the block form is \GthirtytwoBlockElem$\times$ the diagonal, and
it closes the gap: $\mathrm{KL}/(\alpha^2 q_\text{block})$ is \GthirtytwoBlockQ{} by the mean and \GthirtytwoBlockQTrim{} trimmed at
$\alpha=1/4$ (\GthirtytwoBlockOne{} and \GthirtytwoBlockOneTrim{} at $\alpha=1$, where a higher-order term adds). The full form adds
nothing beyond the block form: its ratio to it, \GthirtytwoFullBlock, lies inside the range obtained by flipping the sign of each
tensor's projection at random (\GthirtytwoFullBlockLo--\GthirtytwoFullBlockHi), so the missing term is entirely \emph{within} tensors.
The block and full forms have heavier tails, so their trimmed ratios run above their means. On Qwen3-8B, Qwen3-4B and OLMo-2 the same
comparison gives block/diagonal ratios of \GthirtytwoBlockElemSmallLo--\GthirtytwoBlockElemSmallHi{} for every scheme (\GthirtytwoBlockElemInt{}
and \GthirtytwoBlockElemFp{} for int8 and fp8 at 32B): the diagonal and block forms are indistinguishable there, and 32B int4 is the first
cell that separates them. Per sequence, the diagonal form does not rank int4 at 32B at all (Spearman \GthirtytwoSpFour); the block form
ranks it weakly (\GthirtytwoSpBlockFour). The excess sits in one tensor (post hoc). The last layer's down projection carries \GthirtytwoTopExcess\% of
$q_\text{block}-q_\text{elem}$ for int4: its block form is \GthirtytwoTopOwn$\times$ its diagonal, so it holds \GthirtytwoTopElemShare\% of the
diagonal $q$ but \GthirtytwoTopBlockShare\% of the block form. Without it the block/diagonal ratio is \GthirtytwoRestBlockElem, inside the
band of the smaller models, and an estimator that uses the block form for that tensor alone and the diagonal elsewhere gives
$\mathrm{KL}/(\alpha^2q)=\GthirtytwoHybQ$ (\GthirtytwoHybQTrim{} trimmed) at $\alpha=1/4$. The coherent term is spread over sequences (the top
5\% carry \GthirtytwoTopSeqShare\% of its $d_\ell^2$) with no preferred sign (mean sign \GthirtytwoTopSign); int8 rounding of the same tensor
shows none (\GthirtytwoTopOwnInt), nor do the last-layer down projections of Qwen3-8B, Qwen3-4B and OLMo-2
(\LastDownOwnSmallLo--\LastDownOwnSmallHi). We expected the cause in the weights---an int4 scale stretched by outliers rounds many weights
to zero, so $\Delta$ shrinks $w$ and $d_\ell$ picks up the derivative along $w$---and wrote that prediction down before computing the
weight statistics. It failed: the tensor's correlation between $\Delta$ and $w$ is \CohLastCorr{} (rank \CohLastCorrRank{} of \CohN{} in
the shrinking direction), its zero-code share \CohLastZero{} and its median group max/median \CohLastOutlier, close to Qwen3-8B's last
down projection (\CohEightCorr, \CohEightZero, \CohEightOutlier), which is not coherent; the tensors that do shrink, the down
projections of layers 1--4 in both models (correlation $\CohShrunkCorrLo$ to $\CohShrunkCorrHi$, up to \CohShrunkZeroHi\% of codes at zero),
have own ratios of only \CohShrunkOwnLo--\CohShrunkOwnHi. Local statistics do not single it out either: the mean squared $z$-scores of its
row means, column means and per-group correlations with $w$ are \LocalLastLo--\LocalLastHi, as for the controls (\LocalCtrlLo--\LocalCtrlHi),
while a shrinking tensor shows a row bias of \LocalShrunkRow. Yet an error whose signs are independent of the gradient reaches an own
ratio of \GthirtytwoTopOwn{} with probability below $\SignNullP$ whatever the Fisher block (a Chernoff bound for a
Rademacher quadratic form). The excess therefore needs this
tensor's int4 error and its gradient together---an alignment of the rounding pattern with the tensor's Fisher block that neither side
shows alone---and per-tensor projections cannot resolve it.

A pre-registered follow-up (G34, same \GthirtytwoN{} sequences, fp32) quantized that tensor alone, every target except it, and that tensor
with stochastic rounding (two seeds); its gates reproduced the stored projections and int4 KL exactly. The frozen verdict is \emph{\GthirtyfourVerdict}.
The other tensors behave like the smaller models (KL/$q_\text{elem}$ \GthirtyfourRest{} trimmed); the one tensor alone carries
\GthirtyfourOnlyElem$\times$ its diagonal prediction at $\alpha=1/4$ and, at $\alpha=1$, a KL of \GthirtyfourOnlyOne{} nats, as much as the fully quantized
model (\GthirtytwoFourKl), against \GthirtyfourRestOne{} for all the others; the two parts add up (\GthirtyfourAdd{} of the full KL at
$\alpha=1/4$); and stochastic rounding of the same tensor removes the coherence (own ratio \GthirtyfourSrOwnLo{} and \GthirtyfourSrOwnHi). What
failed is the trimmed agreement of these single-tensor arms with their block form (\GthirtyfourOnlyTrim{} and \GthirtyfourSrTrimLo--\GthirtyfourSrTrimHi{}
against a band of $[0.85,1.15]$). Post hoc, their plain means agree (\GthirtyfourOnlyMean, \GthirtyfourSrMeanLo--\GthirtyfourSrMeanHi), and
the trimmed miss follows from the identity of \S\ref{sec:heavy}: one tensor's $d_\ell^2$ is far heavier-tailed than the KL (its top 5\% of
sequences carry \GthirtyfourTailQ\% against \GthirtyfourTailK\%). Also post hoc, per-row and per-column projections locate the coherence: none
within output rows (row-block/diagonal \GthirtyfourRowBlock), strong within input columns (\GthirtyfourColBlock$\times$), and two of the tensor's 25{,}600 input columns carry \GthirtyfourTwoCols\% of the column-block form; summed, their projections account for
\GthirtyfourPairShare\% of the tensor's excess over its diagonal, and they are correlated across sequences at \GthirtyfourPairCorr. The rounding
errors of two input channels of one tensor, aligned along the gradient, are almost all of what the diagonal estimator misses at 32B. Independent per-weight noise thus fails there; one
model does not say how this scales. The size of the KL
(\GthirtytwoFourKl{} nats per sequence against \EightFourKl{} on Qwen3-8B) is a warning here, unlike on OLMo-2.

\section{Determinism has a price, and the noise floor is one number}\label{sec:floor}

Numerical mismatch is only meaningful above the run-to-run noise floor of the serving engine. vLLM's batch-invariant mode removes that
noise by fixing reduction order; on GH200 at 32 concurrent requests it ran at \BatchInvRatio$\times$ the throughput of the default mode.

We measured that floor on GH200 with a frozen criterion (teacher-forced Qwen3-8B, 256 samples of 768 tokens): every comparison in which
the reduction order differs---other kernel selections, two- or four-way tensor parallelism, batch-invariant against default
kernels---gave \FloorReLo--\FloorReHi{} nats per sequence (the \FloorN{} pre-registered comparisons: \FloorLo--\FloorHi), the same
as on H100, while reruns that reuse the kernel selection were bitwise identical. The batch size itself barely matters: reading the same
autotune cache, 8 and 256 concurrent sequences agree on \FloorBatchEq\% of the token log-probabilities (KL \FloorBatchKL). Because
the autotuned kernel selection was shared within the run, the floor is best read as that of an independent bf16 rounding
realisation, set by which kernels run rather than by how requests are batched. KL measured across engine instances or configurations
carries this floor, so differences below about 0.2 nats per sequence between such measurements are not resolvable unless the kernel
selection is fixed; batch-invariant kernels cost \FloorTfBi$\times$ throughput in this teacher-forced setting.

\section{Guidance for a quantization screen}\label{sec:guide}

(i) Use the Fisher screen for ranking configurations on a fixed sample set. (ii) Compute per-sequence gradients in fp32, or check a
bf16 estimate against fp32 per sequence; fp32 avoids catastrophic outliers but not the heavy tail. (iii) Report absolute predictions with
a trimmed average of the per-sequence quadratic form, which landed within $\pm15\%$ in \GfifteenOneOk{} of \GfifteenOneN{} cells on Qwen3-4B
and OLMo-2 and \GthirtytwoOneOk{} of \GthirtytwoOneN{} on Qwen3-32B (every miss is int4). (iv) Check large perturbations such as int4 against a measured KL, or an $\alpha$ scan, on the target model: on OLMo-2
a quartic term made the expansion under-predict int4 by \OlmoFourTrim$\times$, and the size of the KL did not warn of it; on
Qwen3-32B the diagonal form under-predicted int4 by \GthirtytwoFourTrim$\times$, so also record each tensor's projection $d_\ell$ and
compare the diagonal with the tensor-block form $\tfrac12\sum_\ell d_\ell^2$, which costs nothing extra; at 32B one tensor carried
the gap, so the per-tensor comparison also says where to look.

\section{A quantization screener}\label{sec:screener}

The findings so far say what a Fisher screen must do to be trusted: use per-tensor projections rather than the element-wise diagonal
(\S\ref{sec:big}), summarise sequences with a trimmed mean (\S\ref{sec:heavy}), compute gradients in fp32 (\S\ref{sec:bf}), and stay
away from mixture-of-experts routing. We built a screener that does exactly this and nothing more, and tested it, pre-registered, on a
model family that played no part in any of the earlier analysis.

\paragraph{One pass, every format.}
A forward and backward pass per calibration sequence records, for every target tensor $t$ and every candidate format $f$,
$d_1 = \sum_i g_i\Delta_i$ and $d_2 = \sum_i (g_i\Delta_i)^2$, where $g$ is the gradient of the sequence's continuation log-likelihood and
$\Delta = \hat w_f - w$. Each tensor's gradient is consumed by a post-accumulate hook as soon as it is complete and freed, so memory is
one fp32 model plus one gradient tensor; the quantizers are the same functions that later write the deployment checkpoint. After this
pass every format assignment $c$ costs a sum: the tensor-block prediction $q(c) = \tfrac12\sum_t d_1[c_t,t]^2$ per sequence
(\S\ref{sec:theory}), whose 5\%-trimmed mean $\bar q$ is the screener's point prediction. On \ScrModel{} the pass took
\ScrFisherSecPerSeq{}\,s per sequence on one GH200 (\ScrFisherMin{} minutes for \ScrN{} sequences, three formats, \ScrTensors{} tensors).

\begin{figure}[t]
\centering
\includegraphics[width=\columnwidth]{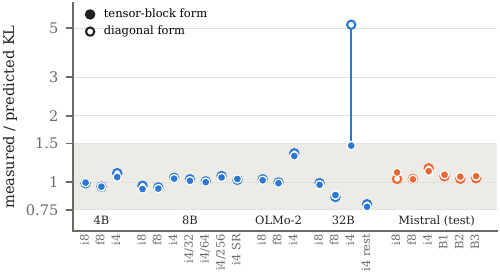}
\caption{Measured over predicted KL (5\%-trimmed) for the \ScrCalibN{} archived dense cells at $\alpha=1$ that set the band (blue)
and the six pre-registered test cells on \ScrModel{} (orange), under the tensor-block form (filled) and the element-diagonal form
(open). Shaded: the frozen band $[\ScrBandLo,\ScrBandHi]$. i4 rest: Qwen3-32B int4 without its one coherent tensor; $B_k$: budgets.}
\label{fig:band}
\end{figure}

\paragraph{A band, not a number.}
The heavy tails of \S\ref{sec:heavy} make the ratio of measured to predicted KL vary from cell to cell even when the expansion holds.
The screener therefore reports $[\ScrBandLo, \ScrBandHi]\times\bar q$: the envelope, rounded outward, of that ratio over all
\ScrCalibN{} archived dense configurations at deployment scale ($\alpha = 1$) that the screener could have produced
(\ScrCalibUniLo--\ScrCalibUniHi{} for the uniform cells of four models; \ScrGRestAlphaOne{} for Qwen3-32B int4 with its one coherent
tensor left unquantized). The band was frozen after that targeted batch and before any measurement on the new model.
Figure~\ref{fig:band} shows these cells under both forms: the diagonal form leaves the band once, at Qwen3-32B int4.

\paragraph{Flag the aligned tensors.}
\S\ref{sec:big} found one tensor whose int4 error is aligned with its own Fisher block. The screener computes each (tensor, format)
pair's own ratio $\overline{d_1^2}/\overline{d_2}$, which is one in expectation when the error's signs are unrelated to the gradient,
and flags pairs with ratio $\ge \ScrTauOwn$ that also carry $\ge \ScrTauExcPct\%$ of the model's predicted KL. Over the
\ScrRetroCells{} archived (model, format) cells, mixture-of-experts included, the rule fires exactly once, on that tensor; the largest
own ratio anywhere else is \ScrRetroMaxOwnOther. A flagged pair is never assigned. A pre-registered batch on that tensor (verdict \ScrGVerdict) shows why the block
form alone is not enough there. Quantized alone to int4 at $\alpha = 1$, the tensor gives a KL of \ScrGOneKlAlphaOne{} nats per sequence,
more than the whole int4 model (\ScrGFullKlAlphaOne); all other tensors together give \ScrGRestKlAlphaOne. Its KL matches its block term
in the mean at $\alpha = \tfrac14$ (\ScrGOneMeanAlphaQuarter) but not at deployment scale (\ScrGOneMeanAlphaOne$\times$ in the mean at
$\alpha = 1$), and the frozen trimmed criteria failed (\ScrGOneAlphaQuarter$\times$ at $\alpha = \tfrac14$) because one tensor's $d_1^2$
is far heavier-tailed than its KL---which is why the screener trims only whole-model sums, never a single tensor's. So the one place
where the diagonal form fails is also where the expansion is least accurate. The alignment sits in the input channels: two of 25,600
columns carry \ScrGTwoColShare\% of the tensor's column-block term. Stochastic rounding at the same four bits removes it (own ratio below
one) and cuts that tensor's KL by \ScrGSrDrop$\times$ at $\alpha = \tfrac14$, with a block prediction that holds in the mean
(\ScrGSrMeanLo--\ScrGSrMeanHi) though not under the frozen trimmed band (\ScrGSrRatioLo--\ScrGSrRatioHi); the screener's ladder does not
include it yet.

\paragraph{Allocate under a budget.}
Given a KL budget $B$, the screener assigns each decoder layer a rung of the ladder int4 (group 128) $<$ int8 (per channel) $<$ bf16.
It starts with every layer at int4 and repeatedly upgrades the layer with the largest reduction of mean predicted KL per added byte
until the band's upper edge, $\ScrBandHi\,\bar q$, is within $B$. Because the block form is additive over tensors, each candidate is
evaluated exactly from the stored projections; the trimmed mean is recomputed from per-sequence sums after every step. The unit of
allocation follows the serving engine: vLLM fuses $q/k/v$ and gate/up projections and requires one scheme per fused layer, so a flag
lifts its whole unit. The output is a compressed-tensors checkpoint with one configuration group per format and exact module names; at
build time every module is decoded by two independent readers and compared bit for bit with the screener's $\hat w$.

\paragraph{Predict throughput.}
Per-token time is modelled as additive over linear layers, so a mixed checkpoint's throughput is the parameter-share harmonic
interpolation of three measured uniform ones (bf16, int8, int4), $1/\mathrm{tps}(c) = \sum_f s_f(c)/\mathrm{tps}_f$. The prediction
for each mixed configuration was written to disk before its checkpoint existed.

\begin{table}[t]
\centering\small
\setlength{\tabcolsep}{3pt}
\begin{tabular}{@{}lrrrrrr@{}}
\toprule
cell & bits & $\bar q$ & KL & KL$/\bar q$ & tok/s & LAMBADA \\
\midrule
int8 & \ScrBitsUEight & \ScrPredUEight & \ScrKlUEight & \ScrRatioUEight & \ScrTpsUEight & \ScrLamUEight \\
fp8 & \ScrBitsUFp & \ScrPredUFp & \ScrKlUFp & \ScrRatioUFp & -- & -- \\
int4 & \ScrBitsUFour & \ScrPredUFour & \ScrKlUFour & \ScrRatioUFour & \ScrTpsUFour & \ScrLamUFour \\
$B_1 = \tfrac12$ & \ScrBitsMOne & \ScrPredMOne & \ScrKlMOne & \ScrRatioMOne & \ScrTpsMOne{} (\ScrTpsPredMOne) & \ScrLamMOne \\
$B_2 = \tfrac14$ & \ScrBitsMTwo & \ScrPredMTwo & \ScrKlMTwo & \ScrRatioMTwo & \ScrTpsMTwo{} (\ScrTpsPredMTwo) & \ScrLamMTwo \\
$B_3 = \tfrac18$ & \ScrBitsMThree & \ScrPredMThree & \ScrKlMThree & \ScrRatioMThree & \ScrTpsMThree{} (\ScrTpsPredMThree) & \ScrLamMThree \\
bf16 & 16.00 & -- & -- & -- & \ScrTpsBf & \ScrLamBf \\
\bottomrule
\end{tabular}
\caption{Pre-registered screener test on \ScrModel{} (unseen family). bits: weight bits per target parameter; $\bar q$ and KL:
trimmed means (nats per 256-token sequence) of predicted and exact fp32 KL; budgets as a fraction of the int4 prediction; tok/s:
vLLM, 128 requests $\times$ 256 tokens, 32 concurrent (predicted in parentheses). Band $[\ScrBandLo,\ScrBandHi]$.}
\label{tab:screener}
\end{table}

\paragraph{Pre-registered test on an unseen family.}
We froze the band, the flag rule, the ladder, three budgets ($\tfrac12$, $\tfrac14$ and $\tfrac18$ of the predicted uniform-int4 KL),
the throughput rule and the scorer, then ran them on \ScrModel{} (\ScrParamsB{}\,B target parameters, \ScrTensors{} tensors), a family
never used in this paper, with \ScrN{} calibration sequences. Three criteria: the band contains the exact KL in at least five of six cells
(three uniform formats, three budgets); each budget is met by the measured KL; each mixed configuration's vLLM throughput is within 10\%
of its prediction. All three held (Table~\ref{tab:screener}; verdict \ScrVerdict). The band contained all \ScrHits{} cells, and the
measured ratios were much tighter than the band, \ScrRatioLo--\ScrRatioHi{} (post hoc: \ScrHitsDraft/6 also under the drafted lower edge
0.85, and \ScrHitsGfifteen/6 under the $\pm15\%$ band of \S\ref{sec:trim}, where the diagonal form gets \ScrHitsElemGfifteen/6 because
uniform int4 sits at \ScrRatioElemUFourX, a difference within noise); the three budgets (\ScrBudgetMOne, \ScrBudgetMTwo{} and
\ScrBudgetMThree{} nats) were met with a third to spare, because the allocator plans against the band's upper edge; and the measured
throughput of the mixed checkpoints was within \ScrTpsErrMax\% of the prediction. The 10\% tolerance did not discriminate at this
operating point: all three uniform throughputs lie within 10\% of the mixed ones, so predicting the nearest uniform format (int8) would
also have passed. The informative comparison is post hoc: that baseline is off by \ScrTpsBaseEightLo--\ScrTpsBaseEightHi\% and bf16 by
\ScrTpsBaseBfLo--\ScrTpsBaseBfHi\%, against the formula's \ScrTpsErrMax\%, which is the repeat-to-repeat spread (\ScrTpsRepSpread\%). The allocation is not uniform in depth: at the loosest
budget the screener kept \ScrMOneIntFourLayers{} of \ScrLayers{} layers at int4, nine of them a contiguous block in the upper-middle of
the network (layers 21--29), and moved the rest, including the first and last layers, to int8.

Two things the test did \emph{not} exercise. The flag never fired on \ScrModel{} (largest own ratio \ScrMaxOwnFour{} under int4), and
with no aligned tensor the diagonal form puts \ScrHitsElem{} of the six cells in the band too; on this model the screener's advantage
over the diagonal form is a guard against the Qwen3-32B case, not a measured gain. And the absolute gains from weight-only formats are
modest at this operating point (512-token prompts, 32 concurrent requests): int4 serves \ScrGainUFour\% more tokens per second than bf16,
int8 \ScrGainUEight\%, and the three budgets \ScrGainMOne, \ScrGainMTwo{} and \ScrGainMThree\%.

\paragraph{Against a diagonal allocator.}
What the block form and the flag buy over a Hessian-trace-style allocator can be read off the stored projections (post hoc). We ran
the same greedy allocator on the element-diagonal form, without flags, under budgets of $\tfrac12$, $\tfrac14$ and $\tfrac18$ of the
diagonal uniform-int4 prediction, a yardstick both can read. On \ScrModel{} the two choose the same format for
\ScrAcMisAgreeLo--\ScrAcMisAgreeHi\% of tensors at the same bits. On Qwen3-32B---in sample for the flag, so an illustration rather
than a test---they agree on \ScrAcQAgreeLo--\ScrAcQAgreeHi\%, and both lift the coherent tensor's layer to int8, because that layer
is costly even on the diagonal. They part only where the budget is loose: from $\ScrBandHi\times\ScrAcQElemUFour \approx
\ScrAcQCross$ nats the diagonal allocator accepts uniform int4, whose measured KL is \GthirtytwoFourTrim$\times$ its prediction
(\S\ref{sec:big}), about \ScrAcQOver$\times$ that budget; at the same budget the screener moves that one tensor to int8 and nothing
else, for \ScrAcQBitsDelta{} more bits per parameter (\ScrAcQBitsBlk{} against \ScrAcQBitsElm), and predicts \ScrAcQBlkPred{} nats;
with all other tensors at int4 and that one in bf16, the measured mean KL was \ScrGRestKlAlphaOne. The two allocators thus differ only
in the regime where a diagonal screen recommends int4, and there by one tensor.

\paragraph{The calibration set must not be small.}
A one-GPU smoke run with eight calibration sequences (unscored, before the main run) gave measured-to-predicted ratios of
\ScrSmokeRatioLo--\ScrSmokeRatioHi{} for the mixed configurations but not for the uniform ones. The allocator upgrades the layers whose
block terms are largest \emph{on the calibration sequences}, so the layers left at int4 are, by selection, ones whose terms came out low
there: an optimizer's curse. We wrote this hypothesis down before the main run, with the prediction that at \ScrN{} sequences the mixed ratios would still exceed
both uniform integer cells; that prediction failed: they came out at \ScrMixRatioLo--\ScrMixRatioHi, below int8 and int4
(\ScrRatioUEight{} and \ScrRatioUFour). A post-hoc bootstrap---resample the sequences, re-allocate, evaluate each chosen configuration on
the full sample---puts the optimism at \ScrCurseSmokeLo--\ScrCurseSmokeHi$\times$ with eight sequences, matching the smoke run, and at
\ScrCurseLo--\ScrCurseHi$\times$ with \ScrN. A screener that allocates on its
own calibration set needs a few hundred sequences, or a held-out split for the final prediction.

\paragraph{Task metric and engine KL.}
LAMBADA accuracy (\ScrLamN{} items) ranged from \ScrLamUFour\% (int4) to \ScrLamUEight\% (int8), bf16 \ScrLamBf\%; the three budgets
kept \ScrLamMTwo--\ScrLamMThree\%. Inside vLLM, the mean log-likelihood ratio along the samples estimates the serving engine's own KL (bf16 kernels, its
own reference); it was \ScrVklRatioMTwo--\ScrVklRatioUFour{} of the exact fp32 KL for int4 and the budgets.

\paragraph{Cost.}
Screening three formats and every budget took one Fisher pass, \ScrFisherMin{} GPU-minutes; verifying the six cells exactly took another
\ScrKlMin{} minutes, and each vLLM configuration two minutes to build and measure. The whole pre-registered test used \ScrGpuHMain{}
GPU-hours (\ScrGpuHAll{} with the smoke runs).

\section{Limitations}

Four dense models from two families (4B--32B) and one mixture-of-experts model, whose pre-registered test is inconclusive (the
dense control failed) and is reported as an observation; three schemes; one GPU type; sample sets of 256--512 sequences. The screener was tested on one further dense model, from a third family (\ScrModel{}). No tensor was flagged on it and
the element-diagonal form also landed in the band, so the case for the flag and the tensor-block form rests on the archived
Qwen3-32B cell; at 32 concurrent requests the three budgeted configurations gained only \ScrGainMThree--\ScrGainMOne\%
throughput over bf16 (uniform int4 \ScrGainUFour\%), too small for the $\pm10\%$ throughput criterion to discriminate. The localization of the 32B int4 miss to one tensor is post hoc and rests on one
model; a weight-side explanation was refuted; a pre-registered follow-up confirmed the tensor's role and the effect of
stochastic rounding but missed its trimmed band (verdict mixed), and the two-column localization is post hoc. The trimming fraction was chosen on Qwen3-8B and tested on the other three models only. The bf16 error is made in the
lower half of the backward (one model, four sequences); the responsible block varies by sequence, and its link to massive activations
rests on two models. Higher-order terms were measured only for OLMo-2 int4 and Qwen3-8B (set A).  The original experiment's data were not re-analysed.

\section{Related work}

\paragraph{Second-order sensitivity.} The expansion we test underlies pruning from Optimal Brain Damage and Optimal Brain
Surgeon~\cite{lecun1989obd,hassibi1992obs} to WoodFisher~\cite{singh2020woodfisher}, Hessian- and Fisher-guided mixed-precision allocation
(HAQ~\cite{wang2019haq}, Q-BERT~\cite{shen2020qbert}, HAWQ and its successors~\cite{dong2019hawq,dong2020hawqv2,yao2021hawqv3}), and the
layer- and block-wise second-order objectives of post-training quantizers (AdaRound~\cite{nagel2020adaround}, BRECQ~\cite{li2021brecq},
OBC and GPTQ~\cite{frantar2022obc,frantar2023gptq}); SqueezeLLM~\cite{kim2024squeezellm} weights its codebook by a diagonal Fisher. These
methods use the expansion to \emph{choose} a quantization. We test it as a \emph{predictor} of the KL a fixed quantization induces, on
models fixed in advance, and find where its diagonal form fails. The Fisher we estimate is the one along $\Delta$ from sampled continuations,
the natural-gradient metric~\cite{amari1998natural}, not the empirical Fisher whose limits are discussed by~\cite{kunstner2019limitations}.

\paragraph{Outliers.} Activation-aware and outlier-aware methods---LLM.int8()~\cite{dettmers2022llmint8}, SmoothQuant~\cite{xiao2023smoothquant},
AWQ~\cite{lin2024awq}, OWQ~\cite{lee2024owq}, SpQR~\cite{dettmers2023spqr}, OmniQuant~\cite{shao2024omniquant}---and rotation or
vector codebooks (QuIP, QuIP\#, AQLM~\cite{chee2023quip,tseng2024quipsharp,egiazarian2024aqlm}) protect a few channels or spread their
error. Outlier dimensions~\cite{kovaleva2021busters,bondarenko2023quantizable}, massive activations~\cite{sun2024massive} and single
``super weights''~\cite{yu2024superweight} show that a handful of coordinates carry a model's behaviour. Our 32B finding is of this kind
but in the error, not the weights: two input channels of one tensor whose rounding errors align with the gradient, invisible to every
weight statistic we computed.

\paragraph{Mixture of experts and formats.} Sub-bit compression of MoE models~\cite{frantar2024qmoe} reports quality, not the expansion's
validity; we find that routing flips make the response non-analytic at small perturbations. Block-scaled formats~\cite{rouhani2023mx} and
stochastic rounding~\cite{gupta2015limited} enlarge the set a screen has to rank.

\paragraph{Numerical nondeterminism.} Batch-invariant kernels~\cite{tml2025nondeterminism} and studies of numerical nondeterminism in
LLM inference~\cite{yuan2025nondeterminism} address run-to-run variation; we price the deterministic mode and measure the floor it
removes, which every KL comparison between engine runs carries.

\section*{AI Usage Disclosure}
Generative AI tools assisted with experiment design and execution, code development, literature search, data analysis, and translation, drafting, and polishing of the manuscript. The core research ideas originated with a human author. The authors independently checked the manuscript and its results and take responsibility for the accuracy and integrity of the work.

\bibliographystyle{plain}
\bibliography{refs,refs-extra,refs-related}

\onecolumn
\appendix
\section{Ledger of pre-registered batches}\label{app:ledger}
\begin{table}[!htbp]\centering\footnotesize\setlength{\tabcolsep}{4pt}
\caption{Ledger of the pre-registered batches (all on GH200). The outcome is by the frozen rule; post-hoc readings are marked.}\label{tab:ledger}
\begin{tabular}{p{3.4cm}p{1.9cm}p{5.1cm}p{5.6cm}}\toprule
Batch (section) & Model(s) & Frozen prediction / criterion & Outcome \\\midrule
Scale, group size, rounding (\S\ref{sec:rel}) & Qwen3-8B & quadratic in $\alpha$; group sizes ranked; SR/RN ratio & $R$ ratio
  \AlphaLo--\AlphaHi{} over $16\times$; \GroupLo--\GroupHi; \SrKl{} vs.\ \SrPred{} (pass) \\
Reproduction of the earlier pipeline (\S\ref{sec:bf}) & Qwen3-8B & the $4.2\times$ over-prediction reproduces unchanged & $R=\RexpThreeInt,
  \RexpThreeFp, \RexpThreeFour$ (reproduced); same samples give the same numbers \\
Pre-flagged sequences (\S\ref{sec:bf}) & Qwen3-8B & dropping three sequences flagged before the run restores $R$ & $R=\RpreInt, \RpreFp,
  \RpreFour$ (pass) \\
Where the error enters (\S\ref{sec:bf}) & Qwen3-8B & error confined to the weight-gradient products & \HwgradN/4 (fail): it lives in
  the residual-stream gradient and grows toward the input \\
Lower vs.\ upper half in fp32 (\S\ref{sec:bf}) & Qwen3-8B & the error is made below layer 16 & bottom error \GeiRC{} with the lower half in
  fp32 vs.\ \GeiRD{} with the upper half (pass) \\
Trimmed average on new models (\S\ref{sec:trim}) & Qwen3-4B, OLMo-2 & P1 $R_\text{trim}\in[0.85,1.15]$; P2 mean $<$ trim; P3 scheme ratios
  $\pm25\%$; P4 $\le1\%$ blow-ups & P1 \GfifteenOneOk/\GfifteenOneN; P2 \GfifteenTwoOk/\GfifteenTwoN; P3 \GfifteenThreeOk/\GfifteenThreeN;
  P4 \GfifteenFourOk/\GfifteenFourN{} (misses: OLMo-2 int4; Qwen3-4B blow-ups) \\
$\alpha$ scan (\S\ref{sec:trim}) & OLMo-2 & higher-order term vs.\ a Fisher bias along int4 & higher order: quartic share
  \GsixteenQuarticShare\% at $\alpha=1$, second-order coefficient \GsixteenSecond{} (decided) \\
Larger model, with an int4 $\alpha$ scan (\S\ref{sec:big}) & Qwen3-32B & P1--P4 as above; int4 slope in $[1.9,2.1]$, $|C/K_1|<0.05$ &
  P1 \GthirtytwoOneOk/\GthirtytwoOneN{} (int4 \GthirtytwoFourTrim; fail), P2--P4 pass except P3 \GthirtytwoThreeOk/\GthirtytwoThreeN;
  control failed (slope \GthirtytwoSlope); post hoc: tensor-block form \GthirtytwoBlockQ{} (mean) / \GthirtytwoBlockQTrim{} (trim) at $\alpha=1/4$,
  \GthirtytwoTopExcess\% of the gap in one tensor; weight-side cause predicted, refuted \\
Where the 32B miss lives (\S\ref{sec:big}) & Qwen3-32B & rest in band; one tensor at its block form and $\ge10\times$ its diagonal;
  additive; stochastic rounding: own ratio $\le4$, block form in band & \emph{\GthirtyfourVerdict}: \GthirtyfourRest, \GthirtyfourOnlyElem$\times$, \GthirtyfourAdd,
  own \GthirtyfourSrOwnLo/\GthirtyfourSrOwnHi{} hold; trimmed \GthirtyfourOnlyTrim{} and \GthirtyfourSrTrimLo--\GthirtyfourSrTrimHi{} fail (means \GthirtyfourOnlyMean, \GthirtyfourSrMeanLo--\GthirtyfourSrMeanHi; post hoc) \\
Noise floor (\S\ref{sec:floor}) & Qwen3-8B & U: one floor for every reorder; D: reruns bitwise & U \FloorN/\FloorN{} (\FloorLo--\FloorHi{}
  nats); D bitwise (pass); batch-size arms confounded by a shared autotune cache (post hoc) \\
Screener on an unseen family (\S\ref{sec:screener}) & \ScrModel & measured/predicted KL in $[\ScrBandLo,\ScrBandHi]$ in $\ge5$ of
  6 cells; three KL budgets met; vLLM throughput within $\pm10\%$ & \ScrHits/6 (ratios \ScrRatioLo--\ScrRatioHi), budgets 3/3,
  throughput within \ScrTpsErrMax\% (pass); three smoke runs preceded it, one on 8 sequences of this model; frozen after that smoke:
  mixed cells above both uniform int cells at 256 sequences, \ScrMixRatioLo--\ScrMixRatioHi{} vs.\ \ScrRatioUEight, \ScrRatioUFour{}
  (fail); post hoc: bootstrap optimism \ScrCurseLo--\ScrCurseHi, no tensor flagged, the diagonal form also \ScrHitsElem/6, the
  throughput criterion not discriminating \\
\bottomrule\end{tabular}\end{table}

\end{document}